\documentclass{article}

\usepackage[preprint]{colm2026_conference}

\usepackage[T1]{fontenc}
\usepackage[utf8]{inputenc}
\usepackage{microtype}
\usepackage{graphicx}
\usepackage{subcaption}
\usepackage{tabularx}
\usepackage{booktabs}
\usepackage{latexsym}
\usepackage{amsmath}
\usepackage[table]{xcolor}
\usepackage{enumitem}
\usepackage{float}
\usepackage{multirow}
\usepackage{array}
\usepackage{algorithm}
\usepackage{algpseudocode}
\usepackage{amssymb}
\usepackage{mathtools}
\usepackage{amsthm}
\usepackage{url}
\usepackage{hyperref}
\usepackage{siunitx}
\usepackage[capitalize,noabbrev]{cleveref}

\theoremstyle{plain}

\theoremstyle{definition}

\theoremstyle{remark}

\usepackage{longtable}
\usepackage[section]{placeins}

\newcommand{\ph}[1]{\textcolor{black}{#1}}

\newcommand{\SingleRAG}{Single-candidate RAG}

\newcommand{\RtwoVC}{R2VC}

\newcommand{\RtwoVCFour}{R2VC ($k=4$)}
\newcommand{\RtwoVCEight}{R2VC ($k=8$)}
\newcommand{\RtwoVCSixteen}{R2VC ($k=16$)}

\newcommand{\ProgramFCFive}{ProgramFC ($N=5$)}
\newcommand{\FOLK}{FOLK}
\newcommand{\PACAR}{PACAR}
\newcommand{\FIRE}{FIRE}
\newcommand{\VeGraph}{VeGraph ($k=5$)}

\newcommand{\DPGraphCheck}{DP-GraphCheck ($\bar{P}=5$)}
\newcommand{\SelfChecker}{Self-Checker}

\usepackage{todonotes}

\hypersetup{
  colorlinks=true,
  citecolor=blue,
  linkcolor=blue,
  urlcolor=blue,
  pdftitle={R2VC Architecture for Fact-checking using LLMs},
  pdfauthor={Dhruv Dixit, Paritosh Pandey, Snigdha Chaturvedi},
  pdfkeywords={Fact-checking, LLMs, Modular Architecture, FEVER, NLI}
}

\title{R2VC: Modular Fact-Checking with Retrieval, Verification, and Confidence Calibration}

\author{Dhruv Dixit\thanks{Equal contribution.} \\
Department of Electrical and Computer Engineering \\
Stevens Institute of Technology \\
Hoboken, New Jersey, United States
\And
Paritosh Pandey\footnotemark[1] \\
Department of Computer Science \\
University of North Carolina \\
Chapel Hill, North Carolina, United States
}

\begin{document}

\maketitle

\begin{abstract}
Large language models are increasingly used for automated fact checking, but end-to-end prompting often entangles evidence retrieval, reasoning, and uncertainty estimation, making failures difficult to diagnose and confidence difficult to trust. We present R2VC, a modular retrieve, reason, verify, calibrate architecture for evidence-grounded fact checking with citations and abstention. R2VC combines hybrid sparse+dense retrieval over Wikipedia, a supervised fine-tuned and DPO-aligned generator that produces diverse structured verdict candidates, an external NLI cross-encoder for evidence-based candidate selection, and a lightweight sequence-level calibrator for confidence estimation and selective abstention. On FEVER, an 8B backbone with R2VC achieves 13.74\% higher accuracy than baseline. Ablation studies show that verifier-based candidate selection and confidence calibration are the largest contributors to performance. Removing candidate selection drops FEVER accuracy to 76.24\%, while removing calibration nearly doubles the Brier score to 0.161. A manual analysis of 250 errors further shows that retrieval failures, especially wrong-entity evidence, remain the dominant bottleneck. Together, these results show that modular fact-checking pipelines can substantially improve both predictive accuracy and confidence reliability in open-domain verification.
\end{abstract}


\section{Introduction}

Large Language Models (LLMs) are increasingly used as end-to-end fact checkers, yet they remain brittle when evidence is incomplete, noisy, or adversarial. A single forward pass muddles retrieval, reasoning, and calibration, making it difficult to guarantee both verdict correctness and citation faithfulness. Our goal is to turn an LLM into an evidence-grounded fact-checking system that predicts a veracity label, returns minimally sufficient citations, and exposes a calibrated probability of correctness to enable selective prediction and abstention using Wikipedia based benchmarks. Recent retrieval-augmented generation (RAG) systems have shown that adding an explicit retrieval pipeline can substantially improve performance on knowledge-intensive tasks, including fact checking, by grounding generation in external evidence \citep{glass2022re2g}. Entailment (NLI) models are increasingly used as verifiers in fact-checking and attribution-style pipelines, but their performance can degrade under domain shift when moving from standard NLI benchmarks to naturally occurring claims paired with long, noisy evidence documents \citep{kamoi2023wice}.

Recent systems mitigate hallucination via retrieval-augmented prompting, self-revision, or consistency-based aggregation, but these often couple evidence acquisition, verdict generation, and confidence estimation inside an LLM loop, limiting diagnosability and leaving abstention heuristics poorly calibrated. In particular, retrieval mistakes such as wrong-entity evidence can yield confident errors, and disagreement-based signals do not directly enforce entailment or contradiction against cited passages. We therefore propose R2VC, a modular retrieve, reason, verify, calibrate pipeline. 

Our design is motivated by the broader observation that modular verification pipelines remain attractive because they make system behavior more controllable and components easier to optimize in isolation \citep{besta2024graph}. Recent interactive verification frameworks likewise emphasize that evidence acquisition and verification are major cost and latency drivers, making it useful to tune, replace, or budget these modules independently under different deployment constraints \citep{xie2025fire}. At the same time, tightly coupled search and reasoning loops must explicitly guard against retrieval-induced errors, reinforcing the value of clean interfaces between retrieval, generation, and verification \citep{xu2024search}. We evaluate R2VC on Wikipedia-based fact verification, using VitaminC for supervised training and reporting zero-shot and few-shot (3-shot) label accuracy on VitaminC and FEVER. We compare against component ablations that remove SFT and DPO, verifier-based selection, and calibration or abstention, alongside non-LLM and heuristic baselines under a shared evaluation protocol. We also report FEVER and VitaminC label accuracy across multiple LLMs in Table~\ref{tab:fever_models} and analyze 250 annotated errors to characterize dominant failure modes. Figure~\ref{fig:task-io} summarizes the task interface and provides an illustrative example.

\section{Related Work}

\paragraph{Benchmarks and attribution.}
\emph{FEVER} \citep{Thorne18Fever} introduced large-scale claim verification with sentence-level evidence, and \emph{FEVEROUS} \citep{aly2021feverous} extends this setting to include tabular evidence. WICE \citep{kamoi2023wice} frames real-world claim verification as document-level entailment over Wikipedia claims and their cited sources, highlighting that retrieval and long-context evidence selection remain key bottlenecks for entailment-based verifiers. Our primary training source is \emph{VitaminC}, which provides contrastive claim--evidence pairs mined from Wikipedia revisions \citep{schuster2021get}. We also connect to attribution work: AIS formalizes whether model statements are verifiable against identified passages and proposes automatic attribution metrics \citep{rashkin-etal-2023-measuring}.

\paragraph{Retrieval-augmented fact-checking.}
Retrieval-augmented generation (RAG) conditions outputs on passages retrieved from corpora such as Wikipedia \citep{lewis2020retrieval}. Self-RAG tightly interleaves retrieval, generation, and self-critique to improve factuality and citation accuracy \citep{asai2024self}. RARR follows a ``research and revise'' paradigm that edits an initial response using retrieved evidence \citep{gao2023rarr}. Re2G (Retrieve, Rerank, Generate) extends the RAG paradigm with an explicit reranking stage before generation, enabling a multi-stage pipeline that improves evidence selection for downstream generation \citep{glass2022re2g}. Moreover, Re2G uses the reranker to merge candidate sets from retrievers with incomparable scoring functions (e.g., BM25 and dense retrieval), effectively supporting sparse+dense ensembling under a unified ranking signal. 

\paragraph{Modular pipelines vs.\ tightly-coupled search loops.}
Interactive search paradigms interleave query planning with retrieval feedback and introduce explicit mechanisms (e.g., confidence-gated verification/completion) to mitigate cases where retrieved evidence can mislead the model \citep{xu2024search}.
Agent-style fact-checking frameworks similarly couple retrieval and verification decisions to reduce unnecessary searches and improve efficiency \citep{xie2025fire}.
Modular fact-checking pipelines remain operationally attractive because they localize failures, support clean ablations, and enable incremental component upgrades without retraining the full system end-to-end \citep{besta2024graph}.

\paragraph{Consistency, calibration, and alignment.}
Hallucination detection and confidence estimation are closely related in recent work. Consistency-based methods such as SelfCheckGPT \citep{manakul2023selfcheckgpt} and ConFactCheck \citep{gupta2025consistency} use disagreement across stochastic or cross-model generations as a signal of factual unreliability, while calibration work has shown that neural models are often miscalibrated and that post-hoc fixes such as temperature scaling can remain fragile under distribution shift \citep{guo2017calibration,desai-durrett-2020-calibration}. Although LLMs can sometimes self-report correctness, such estimates do not generalize reliably \citep{kadavath2022language}. Our approach combines these lines of work by using multi-sample candidate generation, but grounding each candidate in retrieved evidence and selecting among them with an external NLI verifier rather than self-consistency alone. To support abstention, we train a Sequence Likelihood Calibration (SLC) model over verifier and generation features, and for alignment we pair supervised fine-tuning with Direct Preference Optimization (DPO) \citep{rafailov2023direct}. This also alings with work on aggregated prompting for improved reasoning \citep{arora2022ask}.


\begin{figure}[t]
    \centering
    \includegraphics[width=\linewidth]{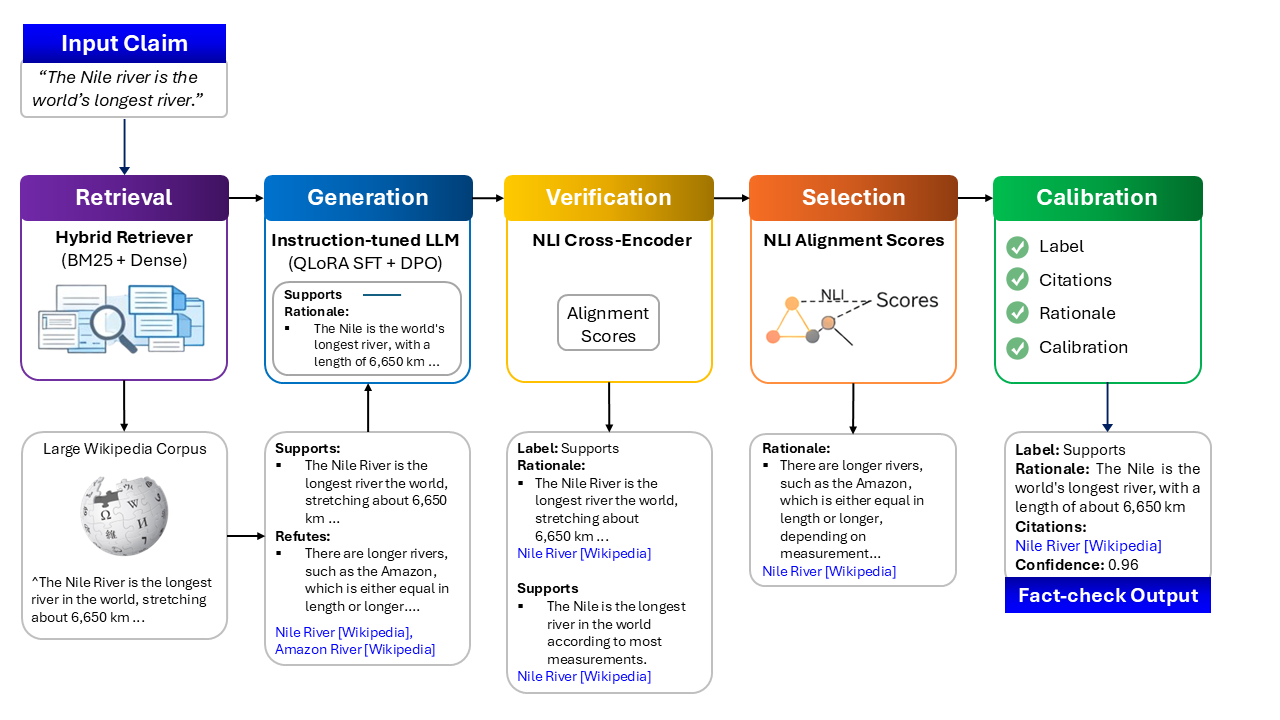}
    \caption{R2VC pipeline: Retrieve $\rightarrow$ Reason $\rightarrow$ Verify $\rightarrow$ Calibrate.}
    \label{fig:architecture}
\end{figure}

\section{Methodology}

\paragraph{\textbf{Overview.}} R2VC (Figure~\ref{fig:architecture}) decomposes fact checking into four stages: hybrid retrieval, candidate generation, external verification, and confidence calibration. Given a claim, the system retrieves evidence from Wikipedia, generates multiple structured candidates, selects among them with an NLI-based verifier, and then applies SLC to produce calibrated confidence and abstention. Algorithmic pseudocode is provided in Appendix~\ref{app:algorithms}.

\begin{figure*}[t]
\centering
\small
\setlength{\fboxsep}{6pt}
\fbox{%
\begin{minipage}{0.97\textwidth}
\textbf{VitaminC example} \\
\textbf{claim:} ``The \emph{Harry Potter and the Goblet of Fire} film was released in 2005.'' \\
\textbf{\texttt{evidence}:} ``\dots \emph{Harry Potter and the Goblet of Fire} is a 2005 fantasy film directed by Mike Newell \dots'' \\
\textbf{label:} \texttt{SUPPORTS}

\vspace{0.7em}
\hrule
\vspace{0.7em}

\textbf{FEVER example} \\
\textbf{claim:} ``The film \emph{Inception} was directed by Christopher Nolan.'' \\
\textbf{\texttt{evidence}:} ``\emph{Inception} is a 2010 science fiction action film written and directed by Christopher Nolan.'' \\
\textbf{label:} \texttt{SUPPORTS}
\end{minipage}%
}
\caption{Illustrative annotated examples from VitaminC and FEVER after standardization into the common claim, evidence, and label format used in our pipeline.}
\label{fig:dataset-examples}
\end{figure*}

\subsection{Datasets}

We work in the standard Wikipedia-based fact-verification setting, using VitaminC for supervised training, FEVER for evaluation, and a large English Wikipedia snapshot as the retrieval corpus. VitaminC~\citep{schuster2021get} is a contrastive benchmark built from Wikipedia revisions in which each example contains a claim, a single evidence sentence, and a label in $\{\text{SUPPORTS}, \text{REFUTES}, \text{NEI}\}$. Many examples are paired with minimally different evidence sentences such that one supports the claim and another refutes it, making the dataset useful for testing sensitivity to subtle factual differences. We use a public Hugging Face version and standardize it into the schema $\{\texttt{claim}, \texttt{evidence\_text}, \texttt{label}\}$, mapping related label variants into the shared ternary space and filtering examples with empty claims. The resulting dataset contains $317{,}672$ training examples, $54{,}012$ validation examples, and $47{,}929$ test examples. Figure~\ref{fig:dataset-examples} shows an illustrative standardized example, while Figure~\ref{fig:vitaminc-claim-evidence-lengths} shows the joint distribution of claim and evidence lengths.

FEVER~\citep{Thorne18Fever} is a large-scale Wikipedia-based fact-verification benchmark with 185,445 claims labeled as \texttt{SUPPORTED}, \texttt{REFUTED}, or \texttt{NOT ENOUGH INFO}, each paired with one or more evidence sentences. We use FEVER as our principal evaluation benchmark and standardize it into the same ternary format as VitaminC by retaining \texttt{NOT ENOUGH INFO} examples and collapsing each evidence set into a flat \texttt{evidence\_text} field through heuristic concatenation. This yields a common representation across training and evaluation data while preserving the task semantics. An illustrative FEVER example is also shown in Figure~\ref{fig:dataset-examples}, and full results across multiple LLMs are reported in Table~\ref{tab:fever_models}. Beyond these labeled datasets, we build a large retrieval corpus from an English Wikipedia snapshot such as November 2023, storing each paragraph as a separate document with fields $(\texttt{doc\_id}, \texttt{title}, \texttt{url}, \texttt{text})$ and indexing it with both Lucene/BM25 and FAISS HNSW. In practice, this produces several million paragraphs, roughly corresponding to the full English Wikipedia article set, all of which are available to the retriever at inference time.

\subsection{Hybrid retrieval}

We use a Wikipedia evidence corpus $\mathcal{D}$ in which each retrieval unit is a paragraph paired with its page title and URL. Let $\mathcal{R}_s$ and $\mathcal{R}_d$ denote the sparse and dense retrievers. For a claim $c$, they return scored lists $E_s=\{(e_j, r^s_j)\}_{j=1}^{N_s}$ and $E_d=\{(e_j, r^d_j)\}_{j=1}^{N_d}$, where $r^s_j$ and $r^d_j$ are the sparse and dense retrieval scores for evidence item $e_j$.

We min--max normalize scores from each retriever independently:
\[
\tilde r^m_j =
\frac{r^m_j - \min_{\ell} r^m_\ell}
{\max_{\ell} r^m_\ell - \min_{\ell} r^m_\ell + \epsilon},
\qquad m\in\{s,d\}.
\]
For each evidence item $e$, we compute the fused retrieval score $r(e \mid c) = \alpha \tilde r^s(e) + (1-\alpha)\tilde r^d(e)$, where $\alpha \in [0,1]$ is the interpolation weight b/w sparse and dense retrieval, and missing scores are treated as $0$. We keep the top-$K$ items under $r(e \mid c)$ as the final evidence pool $\mathcal{E}(c)$.

All later stages operate only on this fused evidence pool, which keeps inference efficient while preserving retrieval recall for downstream generation and verification. In implementation, we fuse results by outer-joining on \texttt{doc\_id}, normalizing scores per retriever, and coalescing passage fields to ensure robust downstream prompting; pseudocode is given in Appendix~\ref{app:algorithms}, and detailed retriever models and hyperparameters are provided in Table~\ref{tab:h100_fullweights_hparams}.

\subsection{Generator training: SFT and DPO}

We do supervised finetuning on VitaminC and preference optimization via DPO on a \texttt{Llama-3.1-8B-Nemotron-Nano}. The training prompt template is provided in Appendix~\ref{app:prompt-template}.

\paragraph{SFT stage.} 
We build an SFT dataset by applying this template to all splits of the standardized VitaminC dataset, using the \texttt{evidence\_text} field as the single evidence snippet. We fine-tune all model parameters without quantization, using bf16 full weights, \texttt{flash\_attention\_2}, maximum sequence length 8192, per-device batch size 2, and gradient accumulation over 8 steps, giving an effective batch size of 16. We use a learning rate of 2e-4, cosine decay with warmup ratio 0.03, weight decay 0.1, the \texttt{paged\_adamw\_8bit} optimizer, gradient checkpointing, and train for 1 epoch. The objective is next-token likelihood over the concatenated prompt and response, with a masking collator that zeroes loss on user tokens and restricts learning to the assistant segment.

\paragraph{Candidate diversity.} For each claim, we first retrieve a fused pool of passages $E(c)$. To encourage diverse candidates without increasing retrieval cost, we generate $k=16$ candidates by pairing random evidence subsets with a fixed bank of diverse decoding configurations. For each $i\in\{1,\dots,k\}$, we sample $S_i \subseteq E(c)$ with $|S_i|=\min(16,|E(c)|)$ and decode $(\hat y_i,\hat r_i)=\mathcal{G}(c,S_i;\theta_i)$, where $\theta_i$ denotes a decoding configuration such as temperature or top-$p$. This produces candidates grounded in the same retrieved evidence pool but exposed to slightly different evidence mixtures and reasoning paths.

\paragraph{DPO stage.} To align the generator toward better citation usage and label correctness, we construct preference triples $(x, y^{+}, y^{-})$ from VitaminC, where $x$ is the fact-checking prompt for a claim-evidence pair, $y^{+}$ is a templated chosen response with the correct True, False, or Uncertain label and a minimal evidence-grounded rationale, and $y^{-}$ is a templated rejected response with either an incorrect label or clearly mismatched citations. We optimize the SFT model with DPO using $\beta = 0.2$ and a learning rate of 4e-6, encouraging the model to assign higher log-probability to $y^{+}$ than to $y^{-}$. This preserves the SFT-induced response format while explicitly discouraging hallucinated or misaligned citations, yielding a final generator $G_{\theta}$ optimized for structured fact-checking outputs conditioned on retrieved evidence.

\paragraph{Candidate Generation.} Given a claim $c$ and fused evidence $\mathcal{E}(c) = \{e_1,\dots,e_K\}$, we generate $k = 16$ candidate answers using a fixed bank of diverse decoding configurations, each conditioned on a random subset of up to 16 evidence passages. For each decode, we parse the raw text into a structured candidate $z$ consisting of a label $z.\texttt{label} \in \{\text{True}, \text{False}, \text{Uncertain}\}$ inferred from the leading token (``True.'', ``False.'', or other), a rationale $z.\texttt{rationale}$ given by the remaining sentence(s), citations $z.\texttt{citations}$ extracted from bracketed citation spans, and the evidence subset $z.\texttt{passages}$ seen by the generator. This produces a candidate set $\mathcal{Z}(c) = \{z_1,\dots,z_k\}$, which is then passed to the verifier for scoring.

\subsection{External verifier (NLI + citation checks)}

To decouple generation from verification, we use a cross-encoder NLI model (\texttt{cross-encoder/nli-deberta-v3-large}) as an external verifier. For each candidate $z$, the verifier evaluates every evidence passage $p \in z.\texttt{passages}$ by scoring the pair (claim, passage) and producing probabilities $(p_{\text{contr}}, p_{\text{neut}}, p_{\text{ent}})$. From these passage-level scores, we derive the maximum entailment support and contradiction signals as $s_{\text{ent}} = \max_{p} p_{\text{ent}}(p)$ and $s_{\text{contr}} = \max_{p} p_{\text{contr}}(p)$. To capture support spread across multiple pieces of evidence, we sort passages by entailment score and average the top-6 values to obtain a coverage term $s_{\text{cov}}$. We also compute a citation-validity score $s_{\text{cit}}\in[0,1]$ as the maximum heuristic similarity between the claim and each candidate passage, using string overlap and numeric matching. Finally, using thresholds $T_{\text{ent}} = 0.60$, $T_{\text{contr}} = 0.60$, and a margin $\delta = 0.10$, we derive an NLI-based label $y_{\text{NLI}} \in \{\text{True},\text{False},\text{Uncertain}\}$ from $(s_{\text{ent}}, s_{\text{contr}})$.

We then compare the generator label $y_{\text{gen}}$ with $y_{\text{NLI}}$ to obtain a label-consistency score $s_{\text{cons}}\in\{0, 0.5, 1\}$, and measure citation alignment through $s_{\text{align}}\in[0,1]$ by checking whether the passages cited by the candidate overlap with those receiving the highest entailment scores. These signals are combined into a scalar verifier score, $s_{\text{ver}}(z) = 0.6\, s_{\text{ent}} + 0.2\, s_{\text{cov}} - 0.3\, \max(0, s_{\text{contr}} - s_{\text{ent}}) + 0.1\, s_{\text{cons}} + 0.1\, s_{\text{align}} - 0.1\, (1 - s_{\text{cit}})$, which rewards strong entailment, broader evidence coverage, and agreement between citations, NLI, and the generator's label, while penalizing excess contradiction and weak citation quality. The verifier also outputs a distribution over NLI labels, $\mathbf{p}_i = \mathcal{V}(c, \hat y_i, S_i) = \big(p_i^{\text{ent}}, p_i^{\text{neu}}, p_i^{\text{con}}\big)$, with $p_i^{\text{ent}}+p_i^{\text{neu}}+p_i^{\text{con}}=1$. For each candidate, we retain both $s_{\text{ver}}(z)$ and its component signals, which are later used for calibration.

\subsection{Sequence Likelihood Calibration (SLC)}

While $s_{\text{ver}}$ is a useful internal score, it is not itself a calibrated probability of correctness. We therefore train a shallow calibrator $f_{\phi}$, implemented as logistic regression, on features extracted from a held-out slice of 10,000 VitaminC examples. These features include the mean log-probability of generated tokens, the verifier score $s_{\text{ver}}$, the agreement fraction across candidates in $\mathcal{Z}(c)$ that share the chosen label, the number of evidence passages used in the candidate, the rationale length in tokens, the label-consistency score $s_{\text{cons}}$, and the citation-alignment score $s_{\text{align}}$. For each training example, we simulate inference with up to 12 generated candidates, select the chosen candidate $z^{\*}$, assign a binary target indicating whether its predicted label matches the gold label, and fit $f_{\phi}$ to predict this target from the resulting feature vector. We evaluate the calibrator using Brier score and expected calibration error (ECE), and store the best checkpoint for use at test time.

\begin{table}[t]
\centering
\captionsetup{font=footnotesize}
\caption{Comparison across models on FEVER and VitaminC. $^\dagger$ denotes Run on Local Machine (RoLM) in the same environment. Calibration metrics (Brier, ECE) are reported only for RoLM runs; for cited baselines, these values were not reported in the original papers and are marked as \textit{n/r}.}
\label{tab:fever_models}
\footnotesize
\setlength{\tabcolsep}{4pt}
\renewcommand{\arraystretch}{1.05}
\resizebox{0.7\textwidth}{!}{%
\begin{tabular}{lccccc}
\toprule
\textbf{Model} & \textbf{Params} & \textbf{Accuracy (\%)} & \textbf{Macro-F1} & \textbf{Brier} & \textbf{ECE} \\
\midrule

\multicolumn{6}{l}{\textbf{FEVER}} \\
\midrule
Mistral-7B \citep{wolfe2024laboratory}                        & 7B   & 69.84             & \textit{n/r} & \textit{n/r} & \textit{n/r} \\
Qwen3-8B$^\dagger$                                           & 8B   & 61.53            & 60.42         & 0.217        & 0.0485       \\
Llama-3.1-8B$^\dagger$                                       & 8B   & 75.14             & 74.87         & 0.192        & 0.0361       \\
Llama-3.1-Nemotron-8B$^\dagger$                              & 8B   & 74.54             & 75.40         & 0.095        & 0.0143       \\
\rowcolor{blue!5}%
Llama-3.1-Nemotron-8B \textbf{+ R2VC}$^\dagger$              & 8B   & 84.71              & 85.68        & 0.083         & 0.0125       \\
\rowcolor{blue!5}%
Qwen3-8B \textbf{+ R2VC}$^\dagger$                           & 8B   & 83.44             & 83.36         & 0.112        & 0.0184       \\
Llama2-13B \citep{anonymous2023factcheckingpdf}               & 13B  & 76.86             & \textit{n/r} & \textit{n/r} & \textit{n/r} \\
Qwen3-14B$^\dagger$                                          & 14B  & 63.65            & 62.54         & 0.196        & 0.0438       \\
Llama2-70B \citep{anonymous2023factcheckingpdf}               & 70B  & 86.4             & \textit{n/r} & \textit{n/r} & \textit{n/r} \\
BLOOM \citep{anonymous2023factcheckingpdf}                    & 176B & 71.0             & \textit{n/r} & \textit{n/r} & \textit{n/r} \\
OpenAI GPT-4 \citep{thibault2025guide}                        & 1.7T & \underline{89.2} & \textit{n/r} & \textit{n/r} & \textit{n/r} \\

\midrule
\multicolumn{6}{l}{\textbf{VitaminC}} \\
\midrule
Qwen3-8B$^\dagger$                                           & 8B   & 88.76            & 88.84         & 0.116        & 0.0315       \\
Llama-3.1-Nemotron-8B$^\dagger$                              & 8B   & 87.81 & 87.84 & 0.083 & 0.0088       \\
\rowcolor{blue!5}%
Llama-3.1-Nemotron-8B \textbf{+ R2VC}$^\dagger$              & 8B   & 99.78             & 99.82         & 0.073        & 0.0077       \\
\rowcolor{blue!5}%
Qwen3-8B \textbf{+ R2VC}$^\dagger$                           & 8B   & 99.21             & 99.15         & 0.0812       & 0.0095       \\

\bottomrule
\end{tabular}%
}
\end{table}

At inference, the calibrated confidence is computed as $p = f_{\phi}(\text{features}(z^{\*}, \mathcal{Z}(c)))$ and we optionally override the predicted label to \texttt{Uncertain} whenever $p < \tau_{\text{cal}} = 0.60$. Together with the earlier verifier-score threshold $\tau_{\text{ver}} = 0.55$, this produces a two-stage abstention mechanism that suppresses predictions when either verification strength or confidence is too low. Concretely, we select the highest-confidence candidate $i^\star=\arg\max_i \pi_i$ and predict
\[
\hat y =
\begin{cases}
\hat y_{i^\star}, & \text{if } \pi_{i^\star}\ge \tau_{\text{cal}} \ \wedge\ s_{\text{ver}}(i^\star)\ge \tau_{\text{ver}},\\
\textsc{Uncertain}, & \text{otherwise.}
\end{cases}
\]
so only those passing both verifier and calibration are returned as final fact-check decisions.

This modular design supports clean ablations of retrieval, alignment, verification, and calibration, allowing component upgrades without retraining full pipeline. We evaluate R2VC against several baselines and ablations, described in the following sections.

\section{Results}
\label{sec:results}

\subsection{Benchmark Performance on FEVER and VitaminC}

Table~\ref{tab:fever_models} reports accuracy, Macro-F1, Brier score, and ECE on FEVER and VitaminC. Entries marked with $^\dagger$ were reproduced locally under a shared evaluation pipeline and prompt template, while the remaining numbers are taken from prior work. Because Brier and ECE are available only for local runs, calibration comparisons are limited to those models.

Within the 8B class, \texttt{Llama-3.1-Nemotron-8B + R2VC} is the strongest locally evaluated model on both benchmarks. On FEVER, it reaches $84.71\%$ accuracy and $85.68$ Macro-F1, compared with $75.14\%$ and $74.87$ for \texttt{Llama-3.1-8B}, and $74.54\%$ and $75.40$ for the base \texttt{Llama-3.1-Nemotron-8B}. Calibration also improves, with Brier dropping from $0.192$ and $0.095$ to $0.083$, and ECE from $0.0361$ and $0.0143$ to $0.0125$. The same trend appears for Qwen: \texttt{Qwen3-8B + R2VC} reaches $83.44\%$ accuracy and $83.36$ Macro-F1 on FEVER, far above the base \texttt{Qwen3-8B} at $61.53\%$ and $60.42$, while improving calibration from $0.217/0.0485$ to $0.112/0.0184$ in Brier/ECE.

R2VC also compares favorably to larger local baselines. On FEVER, \texttt{Qwen3-8B + R2VC} substantially outperforms \texttt{Qwen3-14B}, which attains $63.65\%$ accuracy, $62.54$ Macro-F1, $0.196$ Brier, and $0.0438$ ECE. This indicates that the gains come from the retrieval, verification, and calibration pipeline rather than simply scaling model size. Across both Nemotron and Qwen backbones, lower Brier and ECE confirm that the improvements are not only in classification accuracy but also in how well confidence tracks empirical correctness. Compared with larger open-weight and proprietary systems, \texttt{Llama-3.1-Nemotron-8B + R2VC} remains competitive: on FEVER it trails \texttt{Llama2-70B} ($86.4\%$) by only $1.69$ points and OpenAI GPT-4 ($89.2\%$) by $4.49$ points, while clearly outperforming \texttt{Llama2-13B} ($76.86\%$), \texttt{Mistral-7B} ($69.84\%$), and \texttt{BLOOM} ($71.0\%$).

On VitaminC, the same pattern is even stronger. \texttt{Llama-3.1-Nemotron-8B + R2VC} achieves $99.78\%$ accuracy, $99.82$ Macro-F1, $0.073$ Brier, and $0.0077$ ECE, improving over the base \texttt{Llama-3.1-Nemotron-8B} at $87.81\%$, $87.84$, $0.083$, and $0.0088$, and over \texttt{Qwen3-8B} at $88.76\%$, $88.84$, $0.116$, and $0.0315$. \texttt{Qwen3-8B + R2VC} also performs strongly, reaching $99.21\%$ accuracy, $99.15$ Macro-F1, $0.0812$ Brier, and $0.0095$ ECE. Notably, the base Nemotron model is already better calibrated than the base Qwen model despite slightly lower raw accuracy, but the full R2VC pipeline improves both families further.

These results show that the full R2VC architecture improves both predictive quality and confidence reliability; that too in a fixed budget (8b params). Lower Brier scores indicate more accurate probabilistic predictions, and lower ECE values indicate that predicted confidence is better matched to actual correctness, making the system more reliable for selective prediction and abstention.

\begin{table}[t]
\centering
\captionsetup{font=footnotesize}
\caption{Ablation study of the proposed framework on FEVER and VitaminC using \texttt{Llama-3.1-Nemotron-8B}. Higher is better for Accuracy and Macro-F1, lower is better for Brier and ECE.}
\label{tab:ablations_main}
\footnotesize
\setlength{\tabcolsep}{4pt}
\renewcommand{\arraystretch}{1.08}
\resizebox{\textwidth}{!}{%
\begin{tabular}{lcccccccc}
\toprule
\multirow{2}{*}{\textbf{Variant}}
& \multicolumn{4}{c}{\textbf{FEVER}}
& \multicolumn{4}{c}{\textbf{VitaminC}} \\
\cmidrule(lr){2-5} \cmidrule(lr){6-9}
& \textbf{Acc (\%)} & \textbf{Macro-F1} & \textbf{Brier} & \textbf{ECE}
& \textbf{Acc (\%)} & \textbf{Macro-F1} & \textbf{Brier} & \textbf{ECE} \\
\midrule
\rowcolor{blue!5}%
Full Model (\textbf{R2VC})                              & 84.71 & 85.68 & 0.083 & 0.0125 & 99.78 & 99.82 & 0.073 & 0.0077 \\
w/o DPO                                                 & 83.02 & 83.97 & 0.091 & 0.0138 & 97.78 & 97.82 & 0.080 & 0.0085 \\
w/o Verifier-based Candidate Selection                  & 76.24 & 77.11 & 0.093 & 0.0140 & 89.80 & 89.84 & 0.082 & 0.0086 \\
w/o Confidence Calibration                              & 77.48 & 77.16 & 0.161 & 0.0312 & 96.32 & 96.57 & 0.085 & 0.0227 \\
Single-candidate Generation (no multi-sample selection) & 78.78 & 79.68 & 0.088 & 0.0133 & 92.80 & 92.83 & 0.077 & 0.0082 \\
NLI-only Baseline                                       & 74.54 & 75.40 & 0.095 & 0.0143 & 87.81 & 87.84 & 0.083 & 0.0088 \\
\bottomrule
\end{tabular}%
}
\end{table}

\section{Ablation Study}

Table~\ref{tab:ablations_main} shows that each major component of R2VC contributes to both predictive performance and calibration. The full model performs best on all four metrics across both FEVER and VitaminC, reaching $84.71\%$ accuracy, $85.68$ Macro-F1, $0.083$ Brier, and $0.0125$ ECE on FEVER, and $99.78\%$ accuracy, $99.82$ Macro-F1, $0.073$ Brier, and $0.0077$ ECE on VitaminC. Across ablations, the largest losses come from removing verifier-based candidate selection, confidence calibration, or multi-sample generation, indicating that these are the main drivers of the full system’s gains.

Removing DPO causes a smaller but consistent degradation. On FEVER, accuracy drops from $84.71\%$ to $83.02\%$, Macro-F1 from $85.68$ to $83.97$, Brier worsens from $0.083$ to $0.091$, and ECE from $0.0125$ to $0.0138$. On VitaminC, the same ablation reduces accuracy from $99.78\%$ to $97.78\%$, Macro-F1 from $99.82$ to $97.82$, and calibration from $0.073/0.0077$ to $0.080/0.0085$ in Brier/ECE. This suggests that DPO improves both label correctness and evidence-label alignment, but its effect is smaller than that of selection and calibration.

Removing verifier-based candidate selection causes one of the largest drops in overall quality: FEVER falls to $76.24\%$ accuracy, $77.11$ Macro-F1, $0.093$ Brier, and $0.0140$ ECE, while VitaminC drops to $89.80\%$ accuracy, $89.84$ Macro-F1, $0.082$ Brier, and $0.0086$ ECE. Single-candidate generation also underperforms the full multi-sample system, reaching only $78.78\%$ accuracy, $79.68$ Macro-F1, $0.088$ Brier, and $0.0133$ ECE on FEVER, and $92.80\%$ accuracy, $92.83$ Macro-F1, $0.077$ Brier, and $0.0082$ ECE on VitaminC. Together, these results show that sampling multiple candidates and selecting among them with the verifier is critical for finding stronger evidence-grounded outputs.

Disabling confidence calibration has the clearest effect on reliability. On FEVER, Brier nearly doubles from $0.083$ to $0.161$ and ECE rises from $0.0125$ to $0.0312$, while accuracy and Macro-F1 also drop to $77.48\%$ and $77.16$. On VitaminC, removing calibration reduces performance to $96.32\%$ accuracy, $96.57$ Macro-F1, and worsens calibration from $0.073/0.0077$ to $0.085/0.0227$. These results show that the SLC module is not only improving confidence quality, but also helping final decision quality, especially on FEVER where confidence becomes much less aligned with correctness without calibration.

The NLI-only baseline is clearly weaker than the full pipeline, with $74.54\%$ accuracy and $75.40$ Macro-F1 on FEVER and $87.81\%$ accuracy and $87.84$ Macro-F1 on VitaminC, along with worse Brier and ECE on both datasets. Relative to this baseline, the full model improves FEVER by $+10.17$ accuracy points and $+10.28$ Macro-F1, while reducing Brier from $0.095$ to $0.083$ and ECE from $0.0143$ to $0.0125$; on VitaminC, it improves accuracy by $+11.97$ points and Macro-F1 by $+11.98$, while reducing Brier from $0.083$ to $0.073$ and ECE from $0.0088$ to $0.0077$. These ablations show that the strongest gains come from verifier-based candidate selection, multi-sample generation, and calibration, while DPO provides a smaller but still consistent improvement. More detailed variants are reported in Table~\ref{tab:ablations_appendix} in Appendix~\ref{sec:complete_ablation}.

\begin{figure}[t]
    \centering

    \begin{subfigure}[t]{0.48\linewidth}
        \centering
        \includegraphics[width=\linewidth]{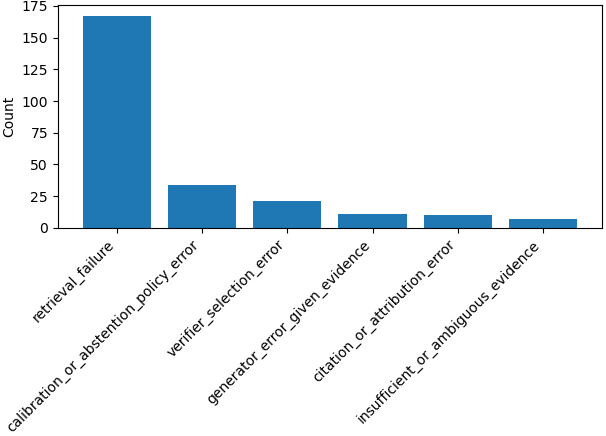}
        \caption{Primary failure categories.}
        \label{fig:error-primary}
    \end{subfigure}
    \hfill
    \begin{subfigure}[t]{0.48\linewidth}
        \centering
        \includegraphics[width=\linewidth]{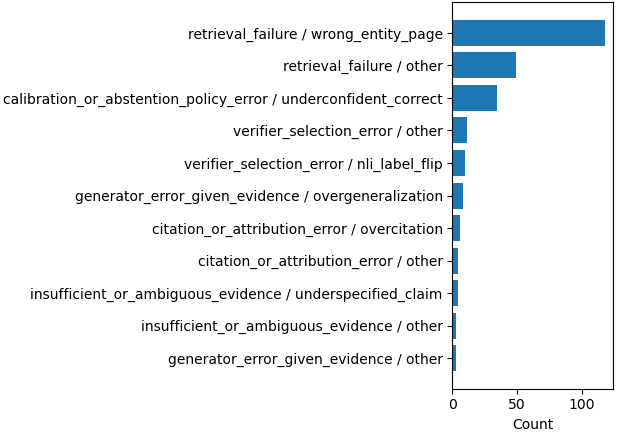}
        \caption{Top error subtypes.}
        \label{fig:error-subtypes}
    \end{subfigure}

    \vspace{0.5em}

    \begin{subfigure}[t]{0.48\linewidth}
        \centering
        \includegraphics[width=\linewidth]{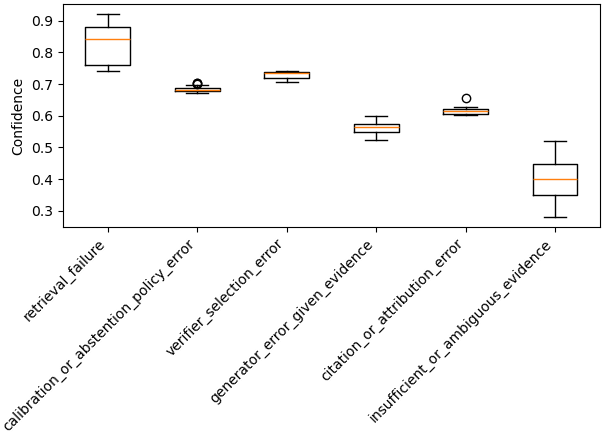}
        \caption{Confidence by primary error category.}
        \label{fig:err-conf-box}
    \end{subfigure}
    \hfill
    \begin{subfigure}[t]{0.48\linewidth}
        \centering
        \includegraphics[width=\linewidth]{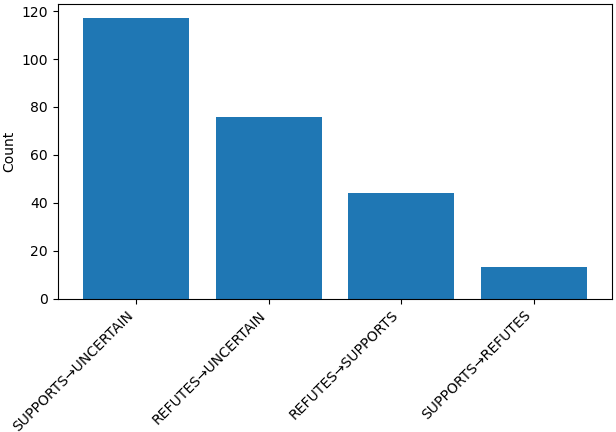}
        \caption{Gold$\rightarrow$predicted label transitions.}
        \label{fig:label-transitions}
    \end{subfigure}

    \caption{Error analysis on the manually annotated error pool. Top left: primary failure categories. Top right: top primary/secondary subtypes, with retrieval errors dominated by wrong-entity evidence and verifier errors often due to NLI label flips. Bottom left: confidence by primary error category, showing overconfidence for retrieval failures and lower confidence for insufficient or ambiguous evidence. Bottom right: gold$\rightarrow$predicted label transitions.}
    \label{fig:error-analysis-combined}
\end{figure}

\section{Failure Modes and Confidence Diagnostics}
\label{sec:error-analysis}

\paragraph{Error sampling and annotation.} We construct an error pool by stratified sampling across (i) calibrated confidence bins, (ii) verifier score bins, (iii) gold label, and (iv) dataset/source. We manually annotated a stratified sample of 250 erroneous predictions using a two-level taxonomy (primary category + secondary subtype), showcased in Appendix~\ref{app:labels-compact} with optional notes. This procedure reduces the chance that the analysis over-represents only low-confidence or single-dataset failures.

Let $\mathcal{D}=\{(c_n, y_n)\}_{n=1}^N$ be a dataset and $\pi_n$ the confidence for the system's non-abstained prediction. In standard benchmark evaluation, \texttt{UNCERTAIN} is scored as the third class, while in selective prediction analysis, only low-confidence threshold-triggered outputs are treated as abstentions. For a threshold $\tau$, define the covered set $\mathcal{C}(\tau)=\{n:\pi_n\ge \tau\}$.
Coverage and risk are
\begin{equation}
\text{Cov}(\tau) = \frac{|\mathcal{C}(\tau)|}{N}, \qquad
\text{Risk}(\tau) = 1 - \frac{1}{|\mathcal{C}(\tau)|}
\sum_{n\in \mathcal{C}(\tau)} \mathbb{I}[\hat y_n = y_n].
\end{equation}
We summarize the trade-off via the area under the risk--coverage curve (AURC).

Figures~\ref{fig:error-primary}--\ref{fig:error-subtypes} report the category distribution and top subtypes. Retrieval errors dominate (Figure~\ref{fig:error-primary}), indicating that evidence acquisition is the primary bottleneck relative to generation, verification, or citation selection. Calibration/abstention errors are the next-largest class, consistent with sensitivity to confidence mapping and decision thresholds. Figure~\ref{fig:error-subtypes} shows that retrieval failures are primarily driven by \emph{wrong\_entity\_page}, suggesting entity disambiguation as a key failure mode. Verifier errors frequently involve \emph{NLI label flips}, where the cross-encoder misclassifies entailment/contradiction and induces incorrect candidate selection. Generator and citation/attribution errors are less frequent but remain relevant for improving rationale faithfulness and citation precision. 

Figure~\ref{fig:err-conf-box} reports calibrated confidence distributions by primary category. Retrieval failures are the most overconfident, consistent with strong internal evidence--claim alignment \emph{conditional on the retrieved set} even when retrieval returns irrelevant or wrong-entity passages. Insufficient/ambiguous-evidence errors occur at the lowest confidence, and many manifest as abstentions (Figure~\ref{fig:label-transitions}). Verifier and calibration/abstention errors concentrate at higher confidence, consistent with confident but incorrect selection or thresholding. Citation/attribution and generator-given-evidence errors occupy intermediate regimes. Figure~\ref{fig:label-transitions} summarizes gold-to-predicted transitions in the annotated error pool. Most errors involve transitions into \texttt{UNCERTAIN}, especially from \texttt{SUPPORTS} or \texttt{REFUTES}, while direct $\texttt{SUPPORTS}\leftrightarrow\texttt{REFUTES}$ polarity flips are less frequent.


\section{Conclusion and Future Work}

\paragraph{Conclusion.} We presented \textbf{R2VC}, a modular fact-checking pipeline that equips an LLM with evidence-grounded verification, citations, and calibrated abstention. By combining hybrid Wikipedia retrieval, SFT+DPO candidate generation, external NLI-based selection, and sequence-level calibration, R2VC achieves strong performance on VitaminC and FEVER while supporting reproducible local evaluation. Our error analysis shows that retrieval remains the main bottleneck, especially when the system retrieves evidence for the wrong entity, leading to confident mistakes. In contrast, insufficient or ambiguous evidence more often results in low confidence and abstention, suggesting that calibration is useful but cannot fully compensate for retrieval failures. Overall, the results highlight retrieval quality, evidence selection, and verifier robustness as the main drivers of further improvement.

\paragraph{Future Work.} A natural next step is to evaluate the same pipeline across a broader range of model families, including reasoning-oriented LLMs, mixture-of-experts architectures, and multimodal models that can use non-textual evidence. We also plan to test R2VC on more challenging fact-checking benchmarks such as DEFAME, and FactLens, and to extend our manual error analysis using the richer taxonomy in Appendix~\ref{app:labels-expanded} to better separate retrieval mismatches, generation and citation failures, verifier ranking errors, and calibration mistakes. This expanded taxonomy can further support targeted training objectives, such as penalizing wrong-entity retrieval, encouraging minimal sufficient citations, and rewarding calibrated abstention when evidence is insufficient, which could be incorporated into future preference optimization or reinforcement learning objectives. By swapping fixed-corpus retrieval module with web-based dynamic evidence retriever, and test the architecture against oen-domain benchmarks like FacTool-QA, BingCheck, and FactCheck-Bench. For these benchmarks, an even more expanded error taxonomy will be required.

\section*{Acknowledgements}
We thank Stevens Institute of Technology for providing GPU resources through the DuckUTE GPU Server, which were used to run all benchmark experiments reported in this paper. We also thank the University of North Carolina at Chapel Hill for providing GPU resources through the Longleaf Cluster, which were used to run the ablation experiments.

\section*{Disclosure of LLM Use}
The authors used an LLM-based writing assistant to revise and improve portions of the manuscript prose based on text originally written by the authors. The authors retained full responsibility for the scientific content of the paper, including the research ideas, technical approach, experiments, results, and conclusions. All such revisions were carefully inspected by the authors for accuracy, and the manuscript was manually checked to ensure that no hallucinated, fabricated, or unsupported content was introduced.

\bibliography{example_paper}
\bibliographystyle{colm2026_conference}

\clearpage
\appendix

\renewcommand{\thefigure}{\thesection.\arabic{figure}}
\renewcommand{\thetable}{\thesection.\arabic{table}}
\renewcommand{\thealgorithm}{\thesection.\arabic{algorithm}}
\renewcommand{\theequation}{\thesection.\arabic{equation}}

\makeatletter
\@addtoreset{figure}{section}
\@addtoreset{table}{section}
\@addtoreset{algorithm}{section}
\@addtoreset{equation}{section}
\makeatother

\section{Fact Checking Illustrative Example}
\begin{figure}[H]
    \centering
    \small
    \setlength{\fboxsep}{6pt}
    \fbox{%
    \begin{minipage}{0.96\linewidth}
    \textbf{Task I/O.}
    \textbf{Input:} claim $c$; evidence corpus $\mathcal{D}$ (Wikipedia paragraphs) indexed by a hybrid retriever (at inference, only $c$ is provided and retrieval over $\mathcal{D}$ is internal). \\
    \textbf{Output:} $(y,E,r,p)$, where $y\in\{\text{True},\text{False},\text{Uncertain}\}$; $E=\{e_1,\dots,e_m\}$ cited passages (doc IDs + offsets); $r$ evidence-grounded rationale; $p\in[0,1]$ calibrated confidence for selective prediction/abstention.

    \vspace{0.35em}
    \hrule
    \vspace{0.35em}

    \textbf{Example.}
    \textbf{Claim:} ``The Great Wall of China can be seen from the Moon with the naked eye.'' \\
    \textbf{Output:} $y=\textbf{False}$;\;
    $E=\{$\emph{Great Wall of China} (Wikipedia)$\}$;\;
    $r$: debunked myth; angular size below naked-eye resolution at lunar distance (with inline citations);\;
    $p=0.87$ (illustrative).
    \end{minipage}}
    \caption{Fact-checking task interface and an illustrative example.}
    \label{fig:task-io}
\end{figure}

\section{Training Prompt Template}
\label{app:prompt-template}

Figure~\ref{fig:prompt-template} shows the prompt--response template used to construct training examples for supervised fine-tuning and preference optimization. We include \texttt{Uncertain} as an allowed output to support abstention when retrieved evidence is missing or conflicting.

\begin{figure}[H]
\centering
\small
\setlength{\fboxsep}{6pt}
\fbox{%
\begin{minipage}{0.96\linewidth}
\textbf{Prompt} \\
\texttt{<|system|>} You are a fact-checking assistant. Use evidence to decide. \texttt{<|user|>} \\
Claim: \{claim\} \\
Candidate evidence: \{evidence\_text\} \\
Output True, False or Uncertain, then a one-sentence justification with citations [Title, \S, line]. \\
\texttt{<|assistant|>}

\vspace{0.5em}
\textbf{Response} \\
\texttt{True. Because \dots [Title \S\ line].} if label = \texttt{SUPPORTS} \\
\texttt{False. Because \dots [Title \S\ line].} if label = \texttt{REFUTES} \\
\texttt{Uncertain. Because \dots [Title \S\ line].} if label = \texttt{NEI}
\end{minipage}%
}
\caption{Prompt--response template used for generator training.}
\label{fig:prompt-template}
\end{figure}

\section{Inference and Retrieval Algorithms}
\label{app:algorithms}

This appendix provides pseudocode for the end-to-end R2VC inference procedure and the sparse-dense retrieval fusion routine used to construct the final evidence pool.

\begin{algorithm}[H]
\caption{Fact checking with fused retrieval, diverse generation, and entailment-based verification}
\label{alg:main}
\begin{algorithmic}[1]
\Require claim $c$; sparse retriever $\mathcal{R}_s$; dense retriever $\mathcal{R}_d$; generator $\mathcal{G}$; entailment verifier $\mathcal{V}$;
top-$N_s$, top-$N_d$, keep-$M$; \#candidates $k$; sample size $s$ (e.g., $\le 6$); thresholds $\tau_{\text{ver}}, \tau_{\text{cal}}$; optional calibrator $g(\cdot)$
\Ensure predicted label $\hat{y}\in\{\textsc{True},\textsc{False},\textsc{Uncertain}\}$ and explanation $r$

\State $E_s \gets \mathcal{R}_s(c, N_s)$ \Comment{BM25 / sparse retrieval}
\State $E_d \gets \mathcal{R}_d(c, N_d)$ \Comment{dense retrieval}
\State $E \gets \textsc{Fuse}(E_s, E_d, M)$ \Comment{Algorithm~\ref{alg:fuse}}

\For{$i \gets 1$ \textbf{to} $k$}
    \State $E_i \gets \textsc{Sample}(E, s)$ \Comment{random subset for diversity}
    \State $\theta_i \gets \textsc{DecodingSchedule}(i)$ \Comment{e.g., temperature/top-$p$ sweep}
    \State $(y_i, r_i) \gets \mathcal{G}(c, E_i; \theta_i)$ \Comment{$y_i \in \{\textsc{True},\textsc{False},\textsc{Uncertain}\}$}
    \State $p_i \gets \mathcal{V}(c, y_i, E_i)$ \Comment{entail/neutral/contradict probs}
    \State $s_i \gets \textsc{VerifierScore}(p_i)$ \Comment{paper's $s_{\text{ver}}$}
    \State $\pi_i \gets
        \begin{cases}
        g(s_i) & \text{if calibrator is used}\\
        \textsc{MonotoneMap}(s_i) & \text{otherwise}
        \end{cases}$
\EndFor

\State $i^\star \gets \arg\max_i \ \pi_i$
\If{$s_{i^\star} < \tau_{\text{ver}}$ \textbf{or} $\pi_{i^\star} < \tau_{\text{cal}}$}
  \State \parbox[t]{\linewidth}{\textbf{return} $(\textsc{Uncertain},$ ``Insufficient / conflicting evidence.'')}
\Else
  \State \textbf{return} $(y_{i^\star}, r_{i^\star})$
\EndIf
\end{algorithmic}
\end{algorithm}

\begin{algorithm}[H]
\caption{Fusion of sparse and dense retrieval results}
\label{alg:fuse}
\begin{algorithmic}[1]
\Require sparse results $E_s$; dense results $E_d$; keep-$M$
\Ensure fused ranked evidence set $E$

\State $E \gets E_s \cup E_d$ \Comment{outer-join / union on document id}

\State \parbox[t]{\linewidth}{%
Normalize sparse scores in $E_s$ to $[0,1]$ via min--max; normalize dense scores in $E_d$ to $[0,1]$.%
}

\For{each passage $e \in E$}
    \State $s_s(e) \gets$ normalized sparse score if present else $0$
    \State $s_d(e) \gets$ normalized dense score if present else $0$
    \State $s(e) \gets \alpha \cdot s_s(e) + (1-\alpha)\cdot s_d(e)$ \Comment{or any fixed fusion rule}
    \State \textsc{CoalesceTextFields}$(e)$ \Comment{ensure valid title/text for prompting}
\EndFor

\State Sort $E$ by $s(e)$ descending and keep top $M$
\State \textbf{return} $E$
\end{algorithmic}
\end{algorithm}

\section{Proposed Expanded Error Taxonomy}
\label{app:labels-expanded}

\begin{table}[H]
\centering
\caption{Proposed expanded taxonomy for finer-grained manual error analysis and subtype-specific training signals.}
\label{tab:labels-expanded}
\scriptsize
\setlength{\tabcolsep}{5pt}
\renewcommand{\arraystretch}{0.95}
\setlength{\extrarowheight}{0pt}
\begin{tabularx}{\textwidth}{@{} >{\raggedright\arraybackslash}p{0.30\textwidth}
                              >{\raggedright\arraybackslash}p{0.20\textwidth}
                              X @{}}
\toprule
\textbf{Primary label} & \textbf{Secondary label} & \textbf{Meaning} \\
\midrule

\multirow[t]{6}{=}{\texttt{retrieval\_failure}} &
\texttt{wrong\_entity\_page} & Retrieved different entity with the same/similar name. \\ \addlinespace[1pt]
& \texttt{lexical\_miss} & BM25 misses paraphrase or rare wording. \\ \addlinespace[1pt]
& \texttt{semantic\_drift} & Dense retrieval returns topical but non-answer passages. \\ \addlinespace[1pt]
& \texttt{multi\_hop\_needed} & Requires chaining; top-$K$ misses the needed hop. \\ \addlinespace[1pt]
& \texttt{timeliness\_mismatch} & Evidence is outdated or claim is time-sensitive. \\ \addlinespace[1pt]
& \texttt{evidence\_buried} & Relevant evidence is in top-$K$ but too low-ranked/unused. \\
\midrule

\multirow[t]{5}{=}{\texttt{insufficient\_or\_ambiguous\_evidence}} &
\texttt{nei\_like} & Claim is not verifiable from the corpus; should abstain. \\ \addlinespace[1pt]
& \texttt{underspecified\_claim} & Missing referent or timeframe. \\ \addlinespace[1pt]
& \texttt{conflicting\_sources} & Retrieved evidence disagrees across sources. \\ \addlinespace[1pt]
& \texttt{requires\_world\_knowledge} & Requires knowledge beyond the provided evidence. \\ \addlinespace[1pt]
& \texttt{requires\_definition} & Term unclear; needs definition/normalization. \\
\midrule

\multirow[t]{7}{=}{\texttt{generator\_error\_given\_evidence}} &
\texttt{negation\_scope} & Misread negation/scope/qualifiers (\emph{not}, \emph{only}, \emph{never}, etc.). \\ \addlinespace[1pt]
& \texttt{numeric\_date\_compare} & Wrong numeric/date reasoning or comparison. \\ \addlinespace[1pt]
& \texttt{entity\_attribute\_swap} & Correct entities retrieved but attributes swapped/misattributed. \\ \addlinespace[1pt]
& \texttt{coreference\_error} & Pronoun/mention refers to the wrong entity. \\ \addlinespace[1pt]
& \texttt{overgeneralization} & Conclusion goes beyond what evidence supports. \\ \addlinespace[1pt]
& \texttt{logical\_composition} & Errors with AND/OR, quantifiers, conditionals, ``at least/most'', etc. \\ \addlinespace[1pt]
& \texttt{hallucinated\_bridge} & Invented intermediate fact not present in evidence. \\
\midrule

\multirow[t]{5}{=}{\texttt{citation\_or\_attribution\_error}} &
\texttt{irrelevant\_citation} & Cited passage does not support the claim/rationale. \\ \addlinespace[1pt]
& \texttt{wrong\_span} & Right document but wrong part/lines cited. \\ \addlinespace[1pt]
& \texttt{missing\_citation} & Key statement lacks supporting citation. \\ \addlinespace[1pt]
& \texttt{overcitation} & Many citations but none directly entail the decision. \\ \addlinespace[1pt]
& \texttt{misquoted\_paraphrase} & Paraphrase distorts evidence meaning. \\
\midrule

\multirow[t]{4}{=}{\texttt{verifier\_selection\_error}} &
\texttt{lexical\_overlap\_bias} & Selects candidate via overlap rather than entailment. \\ \addlinespace[1pt]
& \texttt{nli\_label\_flip} & Verifier misclassifies entailment vs.\ contradiction. \\ \addlinespace[1pt]
& \texttt{ranking\_bug} & Correct candidate exists but scorer prefers the wrong one. \\ \addlinespace[1pt]
& \texttt{evidence\_subset\_issue} & Correct candidate uses better evidence subset but is not selected. \\
\midrule

\multirow[t]{5}{=}{\texttt{calibration\_\allowbreak or\_\allowbreak abstention\_\allowbreak policy\_\allowbreak error}} &
\texttt{overconfident\_wrong} & High confidence despite wrong label/citations. \\ \addlinespace[1pt]
& \texttt{underconfident\_correct} & Low confidence (or abstains) despite being correct. \\ \addlinespace[1pt]
& \texttt{threshold\_too\_low} & Abstention threshold allows too many risky decisions. \\ \addlinespace[1pt]
& \texttt{threshold\_too\_high} & Abstention threshold rejects too many correct decisions. \\ \addlinespace[1pt]
& \texttt{uncertain\_misuse} & Chooses SUPPORTS/REFUTES when it should abstain, or vice versa. \\
\midrule

\end{tabularx}
\end{table}

\FloatBarrier

\section{Error Labels Used in This Work}
\label{app:labels-compact}

\begin{table}[H]
\centering
\caption{Compact error taxonomy used for manual annotation in this work (primary category with coarse subtypes).}
\label{tab:labels-compact}
\scriptsize
\setlength{\tabcolsep}{5pt}
\renewcommand{\arraystretch}{0.95}
\setlength{\extrarowheight}{0pt}
\begin{tabularx}{\textwidth}{@{} >{\raggedright\arraybackslash}p{0.34\textwidth}
                              >{\raggedright\arraybackslash}p{0.20\textwidth}
                              X @{}}
\toprule
\textbf{Primary label} & \textbf{Secondary label(s)} & \textbf{Meaning} \\
\midrule
\texttt{retrieval\_failure} &
\texttt{wrong\_entity\_page}; \texttt{other} &
Evidence retrieval fails, most commonly by retrieving passages about a different but similarly named entity; \texttt{other} covers remaining retrieval issues not otherwise specified. \\

\texttt{insufficient\_or\_ambiguous\_evidence} &
\texttt{underspecified\_claim}; \texttt{other} &
Retrieved evidence is insufficient or the claim is underspecified (e.g., missing referent/timeframe), so a definitive support/refute decision is not justified. \\

\texttt{citation\_or\_attribution\_error} &
\texttt{overcitation}; \texttt{other} &
The decision may be plausible, but citations are not minimally sufficient (e.g., too many or weakly related citations), reducing attribution precision. \\

\texttt{verifier\_selection\_error} &
\texttt{nli\_label\_flip}; \texttt{other} &
The verifier selects the wrong candidate due to entailment/contradiction misclassification (NLI label flip) or other ranking/selection failures. \\

\texttt{calibration\_or\_abstention\_policy\_error} &
\texttt{underconfident\_correct}; \texttt{other} &
Confidence mapping or abstention threshold is misaligned, e.g., the system is insufficiently confident (or abstains) despite being correct. \\
\bottomrule
\end{tabularx}
\end{table}

\FloatBarrier

\section{Additional Figures}
\label{app:additional-figures}

\subsection{Dataset Distribution Plots}

\newcommand{\vitamincfigwidth}{0.45\textwidth}

\begin{figure}[H]
  \centering
  \includegraphics[width=\vitamincfigwidth]{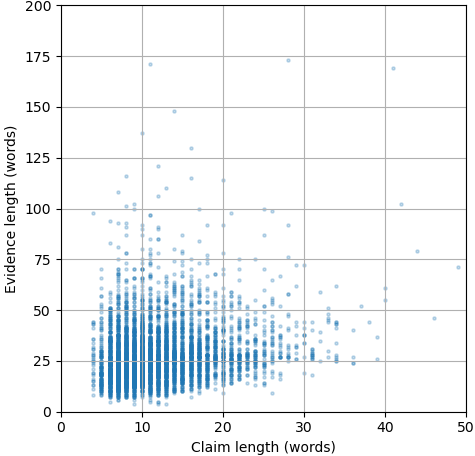}
  \caption{Claim vs.\ evidence length (in words) for a random subsample of VitaminC examples.}
  \label{fig:vitaminc-claim-evidence-lengths}
\end{figure}

\FloatBarrier

\subsection{Training Dynamics Plots}

\begin{figure}[H]
  \centering
  \begin{subfigure}[t]{0.49\textwidth}
    \centering
    \includegraphics[width=\textwidth]{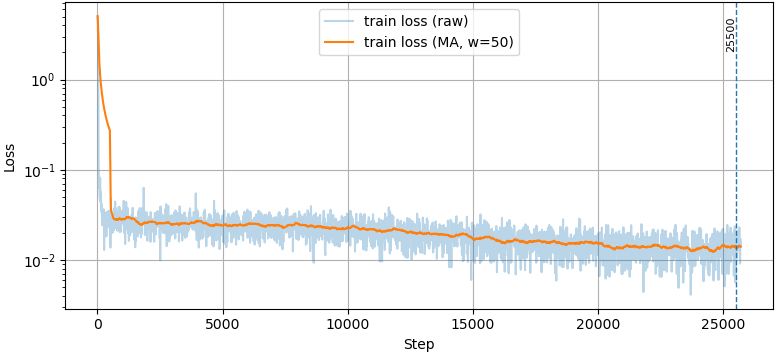}
    \caption{SFT training loss vs.\ step.}
    \label{fig:sft-loss}
  \end{subfigure}\hfill
  \begin{subfigure}[t]{0.49\textwidth}
    \centering
    \includegraphics[width=\textwidth]{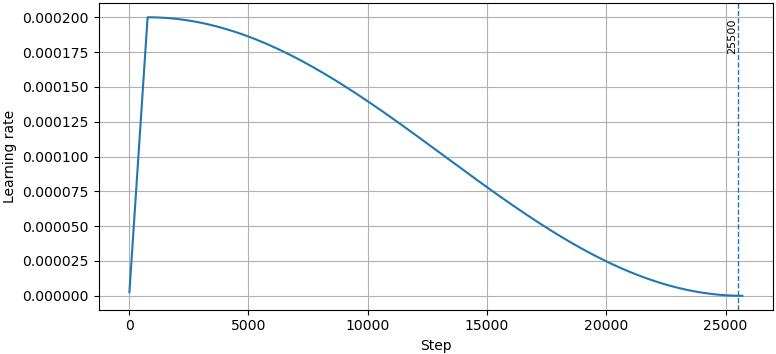}
    \caption{SFT learning-rate schedule.}
    \label{fig:sft-lr}
  \end{subfigure}
  \caption{SFT training dynamics.}
  \label{fig:sft-dynamics}
\end{figure}

SFT training is stable and convergent: Figure~\ref{fig:sft-loss} shows a sharp early loss reduction (roughly two orders of magnitude from an $\mathcal{O}(1)$ initialization into the $10^{-2}$ range) followed by a smooth, steady decline with no signs of divergence or late-stage overfitting, while Figure~\ref{fig:sft-lr} confirms this behavior under a warmup--cosine schedule that ramps to $\approx 2\times 10^{-4}$ by $\sim 10^3$ steps and then decays smoothly toward zero over the remaining $\sim 24$k steps.

\subsubsection{SFT training dynamics}

\FloatBarrier

\subsubsection{DPO training dynamics}

\begin{figure}[H]
  \centering
  \begin{subfigure}[t]{0.49\textwidth}
    \centering
    \includegraphics[width=\textwidth]{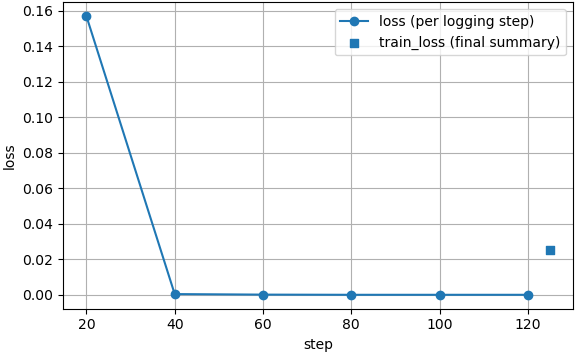}
    \caption{DPO training loss vs.\ step.}
    \label{fig:dpo-loss}
  \end{subfigure}\hfill
  \begin{subfigure}[t]{0.49\textwidth}
    \centering
    \includegraphics[width=\textwidth]{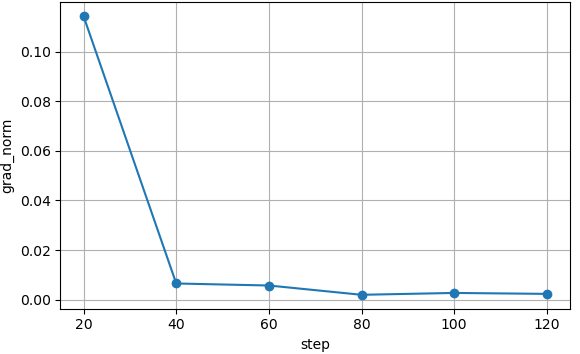}
    \caption{Gradient norm over training.}
    \label{fig:dpo-gradnorm}
  \end{subfigure}
  \caption{Optimization dynamics during DPO.}
  \label{fig:dpo-optim}
\end{figure}

\begin{figure}[H]
  \centering
  \begin{subfigure}[t]{0.49\textwidth}
    \centering
    \includegraphics[width=\textwidth]{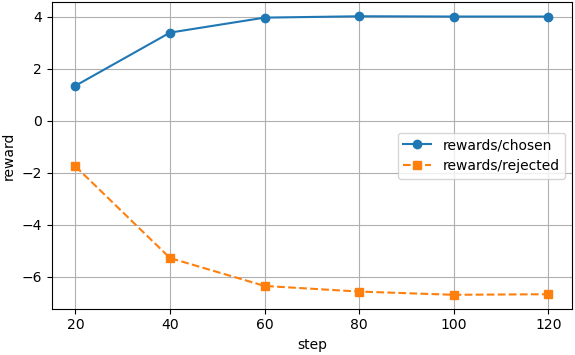}
    \caption{Rewards for chosen vs.\ rejected responses.}
    \label{fig:dpo-rewards}
  \end{subfigure}\hfill
  \begin{subfigure}[t]{0.49\textwidth}
    \centering
    \includegraphics[width=\textwidth]{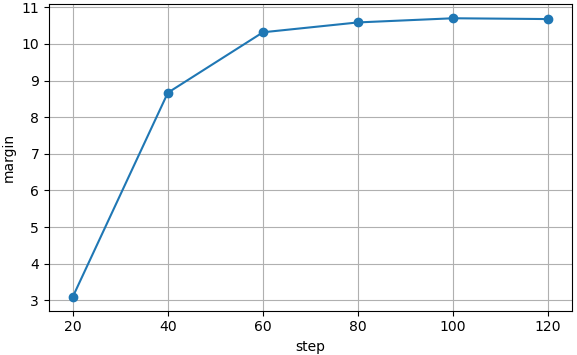}
    \caption{Reward margin over training.}
    \label{fig:dpo-margin}
  \end{subfigure}
  \caption{Preference signal during DPO.}
  \label{fig:dpo-preference}
\end{figure}

\begin{figure}[H]
  \centering
  \begin{subfigure}[t]{0.49\textwidth}
    \centering
    \includegraphics[width=\textwidth]{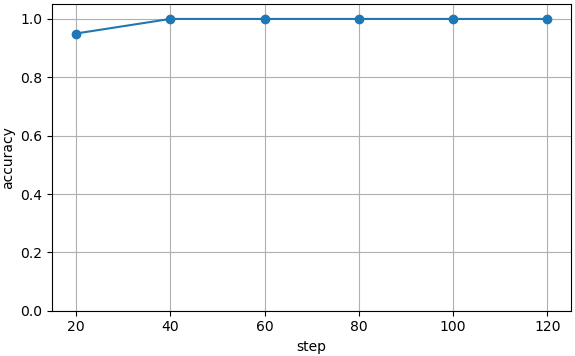}
    \caption{Reward accuracy over training.}
    \label{fig:dpo-accuracy}
  \end{subfigure}\hfill
  \begin{subfigure}[t]{0.49\textwidth}
    \centering
    \includegraphics[width=\textwidth]{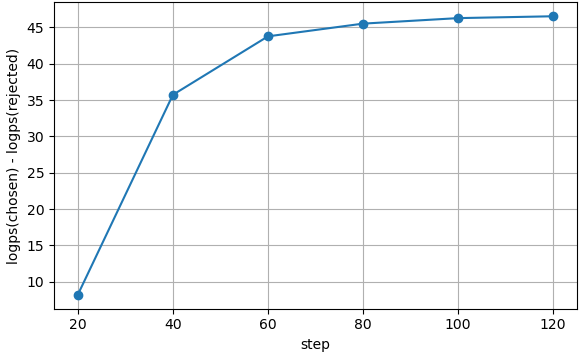}
    \caption{Log-probability margin over training.}
    \label{fig:dpo-logpmargin}
  \end{subfigure}
  \caption{Additional DPO training signals.}
  \label{fig:dpo-signals}
\end{figure}

DPO converges rapidly in a short run: Figure~\ref{fig:dpo-loss} drops from $\approx 0.16$ at step~20 to near-zero by step~40 and stays flat, with the final epoch-level summary still small at step~125 (around $2.5\times 10^{-2}$), indicating strong separation between chosen and rejected responses. This is reflected in reward behavior: Figure~\ref{fig:dpo-rewards} shows chosen rewards increasing from roughly $+1.5$ to about $+4.0$ while rejected rewards decrease from roughly $-1.5$ to below $-6.5$, yielding a widening margin (Figure~\ref{fig:dpo-margin}) from $\sim 3$ to $>10$ and near-perfect preference satisfaction as reward accuracy rises from $\approx 0.95$ to $1.0$ by step~40 and remains at $1.0$ (Figure~\ref{fig:dpo-accuracy}). Optimization signals are consistent with a brief alignment ``polish'' rather than destabilizing updates: Figure~\ref{fig:dpo-gradnorm} shows gradient norms collapsing from roughly $1.1\times 10^{-1}$ to $\mathcal{O}(10^{-3})$, while Figure~\ref{fig:dpo-logpmargin} shows the chosen--rejected log-probability gap increasing from about $8$ to $>45$ nats, all under a low learning rate that begins around $4\times 10^{-6}$ at step~20 and linearly anneals toward zero over $\sim 120$ steps.

\FloatBarrier

\section{Hyper-parameters and environment settings}

\begin{table}[H]
\centering
\caption{Hyperparameter configuration for FEVER run of \texttt{Llama-3.1-Nemotron-8B} on Nvidia H100.}
\label{tab:h100_fullweights_hparams}
\scriptsize
\setlength{\tabcolsep}{5pt}
\renewcommand{\arraystretch}{0.95}
\setlength{\extrarowheight}{0pt}
\begin{tabularx}{\textwidth}{@{} >{\raggedright\arraybackslash}p{0.20\textwidth}
                              >{\raggedright\arraybackslash}p{0.35\textwidth}
                              X @{}}
\toprule
Module & Setting & Value \\
\midrule
Retrieval & Wikipedia corpus cap & No article cap \\
Retrieval & Paragraph cap per article & No paragraph cap \\
Retrieval & Dense retriever & \texttt{BAAI/bge-large-en-v1.5} \\
Retrieval & BM25 top-k & 200 \\
Retrieval & Dense top-k & 200 \\
Retrieval & Fused passages kept & 24 \\
Retrieval & Passages sampled per candidate & 16 \\
Retrieval & FAISS HNSW $M$ & 64 \\
Retrieval & FAISS \texttt{efConstruction} & 512 \\
Retrieval & FAISS \texttt{efSearch} & 512 \\
Retrieval & Dense embedding chunk size & 8192 \\
Retrieval & Dense encoding batch size & 1024 \\
\midrule
Generator / SFT & Finetuning mode & Full-parameter finetuning \\
Generator / SFT & Quantization & None \\
Generator / SFT & Weights / compute dtype & bf16 full weights \\
Generator / SFT & Attention implementation & \texttt{flash\_attention\_2} \\
Generator / SFT & Max sequence length & 8192 \\
Generator / SFT & Per-device train batch size & 2 \\
Generator / SFT & Gradient accumulation & 8 \\
Generator / SFT & Effective SFT batch size & 16 \\
Generator / SFT & Learning rate & 2e-4 \\
Generator / SFT & Weight decay & 0.1 \\
Generator / SFT & Optimizer & \texttt{paged\_adamw\_8bit} \\
Generator / SFT & Gradient checkpointing & on \\
Generator / SFT & Epochs & 1 \\
\midrule
Inference & Rejection-sampling candidates $k$ & 16 \\
Inference & Decoding configurations & 16 temperature and top-p settings \\
Inference & Max new tokens & 160 \\
\midrule
Verifier & NLI cross-encoder & \texttt{cross-encoder/nli-deberta-v3-large} \\
Verifier & Top entailment passages used in scoring & 6 \\
Verifier & Verifier threshold & 0.55 \\
\midrule
Verifier / SLC & Calibration threshold & 0.60 \\
Verifier / SLC & SLC calibration set size & 10{,}000 \\
Verifier / SLC & Candidates per claim during SLC fit & 12 \\
Verifier / SLC & Logistic-regression max iterations & 2000 \\
Verifier / SLC & SLC used at inference & on \\
\midrule
DPO & Preference-pair cap & Full VitaminC train split \\
DPO & Per-device train batch size & 2 \\
DPO & Gradient accumulation & 8 \\
DPO & Effective DPO batch size & 16 \\
DPO & Max target length & 512 \\
DPO & Max sequence length & 8192 \\
DPO & Learning rate & 4e-6 \\
DPO & Weight decay & 0.1 \\
DPO & Optimizer & \texttt{paged\_adamw\_8bit} \\
DPO & DPO $\beta$ & 0.2 \\
DPO & Epochs & 1 \\
\midrule
Runtime & Avg prompt tokens & 608.3 \\
Runtime & Avg generated tokens & 64 \\
Runtime & Generation tokens/sec & 50.92 \\
Runtime & Retrieval latency per claim (s) & 0.0316 \\
Runtime & Generation latency per candidate (s) & 1.2567 \\
Runtime & Verifier latency per candidate (s) & 0.0251 \\
Runtime & Measured full $k$-sample latency per claim (s) & 20.8248 \\
Runtime & Measured full $k$-sample claims/sec & 0.0480 \\
\bottomrule
\end{tabularx}%
\end{table}

\FloatBarrier

\section{Complete Ablation Study}
\label{sec:complete_ablation}

\begin{table}[H]
\centering
\captionsetup{font=footnotesize}
\caption{Incremental and removal ablations of R2VC framework on FEVER and VitaminC. Higher is better for Accuracy and Macro-F1, lower is better for Brier and ECE.}
\label{tab:ablations_appendix}
\footnotesize
\setlength{\tabcolsep}{4pt}
\renewcommand{\arraystretch}{1.08}
\resizebox{\textwidth}{!}{%
\begin{tabular}{lcccccccc}
\toprule
\multirow{2}{*}{\textbf{Variant}}
& \multicolumn{4}{c}{\textbf{FEVER}}
& \multicolumn{4}{c}{\textbf{VitaminC}} \\
\cmidrule(lr){2-5} \cmidrule(lr){6-9}
& \textbf{Acc (\%)} & \textbf{Macro-F1} & \textbf{Brier} & \textbf{ECE}
& \textbf{Acc (\%)} & \textbf{Macro-F1} & \textbf{Brier} & \textbf{ECE} \\
\midrule
\multicolumn{9}{l}{\textit{Incremental ablations from the full model}} \\
\midrule
Base LLM (zero-shot)                                    & 69.47 & 70.27 & 0.119 & 0.0182 & 81.81 & 81.84 & 0.104 & 0.0112 \\
+ SFT                                                   & 71.16 & 71.98 & 0.111 & 0.0169 & 83.81 & 83.84 & 0.097 & 0.0104 \\
+ SFT + DPO                                             & 72.85 & 73.69 & 0.103 & 0.0156 & 85.81 & 85.84 & 0.090 & 0.0096 \\
+ Retrieval-augmented Evidence                          & 74.54 & 75.40 & 0.095 & 0.0143 & 87.81 & 87.84 & 0.083 & 0.0088 \\
+ Multi-candidate Generation                            & 76.24 & 77.11 & 0.093 & 0.0140 & 89.80 & 89.84 & 0.082 & 0.0086 \\
+ Verifier-based Candidate Selection                    & 77.48 & 77.16 & 0.161 & 0.0312 & 96.32 & 96.57 & 0.085 & 0.0227 \\
+ Confidence Calibration                                & 84.71 & 85.68 & 0.083 & 0.0125 & 99.78 & 99.82 & 0.073 & 0.0077 \\
\rowcolor{blue!5}%
Full Model (\textbf{R2VC})                              & 84.71 & 85.68 & 0.083 & 0.0125 & 99.78 & 99.82 & 0.073 & 0.0077 \\

\midrule
\multicolumn{9}{l}{\textit{Removal ablations from the full model}} \\
\midrule
\rowcolor{blue!5}%
Full Model (\textbf{R2VC})                              & 84.71 & 85.68 & 0.083 & 0.0125 & 99.78 & 99.82 & 0.073 & 0.0077 \\
w/o DPO                                                 & 83.02 & 83.97 & 0.091 & 0.0138 & 97.78 & 97.82 & 0.080 & 0.0085 \\
w/o Verifier-based Candidate Selection                  & 76.24 & 77.11 & 0.093 & 0.0140 & 89.80 & 89.84 & 0.082 & 0.0086 \\
w/o Confidence Calibration                              & 77.48 & 77.16 & 0.161 & 0.0312 & 96.32 & 96.57 & 0.085 & 0.0227 \\
Single-candidate Generation (no multi-sample selection) & 78.78 & 79.68 & 0.088 & 0.0133 & 92.80 & 92.83 & 0.077 & 0.0082 \\
NLI-only Baseline                                       & 74.54 & 75.40 & 0.095 & 0.0143 & 87.81 & 87.84 & 0.083 & 0.0088 \\

\bottomrule
\end{tabular}%
}
\end{table}

\section{Rebuttal Stage-1}

\begin{table}[t]
\centering
\small
\caption{Controlled comparison under the evaluation protocol using \texttt{Llama-3.1-Nemotron-8B}. All
retrieval-augmented rows use the same hybrid retriever and generator checkpoint.
Only the downstream selection and calibration mechanisms differ.}
\label{tab:corrected_core}
\resizebox{\linewidth}{!}{
\begin{tabular}{lrrrrrrrr}
\toprule
& \multicolumn{4}{c}{FEVER} & \multicolumn{4}{c}{VitaminC} \\
\cmidrule(lr){2-5} \cmidrule(lr){6-9}
Method & Acc. & Macro-F1 & Brier & ECE & Acc. & Macro-F1 & Brier & ECE \\
\midrule
Claim-only LM
& 61.50 & 71.23 & 0.13 & 0.02
& 84.20 & 79.67 & 0.13 & 0.03 \\

Single-candidate RAG
& 74.54 & 75.69 & 0.12 & 0.03
& 86.83 & 85.40 & 0.11 & 0.02 \\

RAG + 16 samples + majority vote
& 78.11 & 81.24 & 0.12 & 0.03
& 90.45 & 94.47 & 0.12 & 0.03 \\

RAG + 16 samples + generation-logprob selection
& 75.87 & 78.44 & 0.13 & 0.03
& 87.00 & 88.27 & 0.11 & 0.01 \\

RAG + 16 samples + NLI verifier selection
& 80.30 & 82.06 & 0.13 & 0.02
& 96.28 & 94.00 & 0.10 & 0.03 \\
Full R2VC + SLC abstention
& 84.71 & 85.68 & 0.10 & 0.01
& 97.51 & 98.23 & 0.08 & 0.01 \\

Oracle-document R2VC diagnostic
& 86.80 & 91.24 & 0.06 & 0.01
& 97.39 & 97.06 & 0.06 & 0.01 \\
\bottomrule
\end{tabular}
}
\end{table}

\begin{table}[t]
\centering
\small
\caption{Retrieval ablations using \texttt{Llama-3.1-Nemotron-8B}. Evidence recall is
computed against annotated gold evidence where available.}
\label{tab:retrieval_ablation}
\begin{tabular}{lrrr}
\toprule
Retriever & Recall@24 & FEVER Acc. & Retrieval latency \\
\midrule
BM25 only
& 78.91 & 80.61 & 0.02 s \\
Dense only
& 74.01 & 81.29 & 0.03 s \\
Hybrid sparse+dense
& 83.18 & 84.71 & 0.08 s \\
Hybrid + title/entity reranking
& 83.44 & 82.02 & 0.06 s \\
Oracle gold documents
& 100.0 & 86.80 & -- \\
\bottomrule
\end{tabular}
\end{table}

\begin{table}[t]
\centering
\small
\caption{Accuracy-latency trade-off as the candidate budget changes using \texttt{Llama-3.1-Nemotron-8B}.}
\label{tab:k_sweep}
\begin{tabular}{rrrrrr}
\toprule
Candidates $k$ & Acc. & Macro-F1 & Brier & ECE & Latency / claim \\
\midrule
1  & 78.10 & 79.59 & 0.11 & 0.01 & 1.49 s \\
2  & 76.95 & 79.14 & 0.11 & 0.04 & 2.78 s \\
4  & 80.26 & 81.72 & 0.10 & 0.02 & 5.30 s \\
8  & 79.68 & 82.52 & 0.11 & 0.02 & 10.44 s \\
16 & 84.71 & 85.68 & 0.10 & 0.01 & 20.31 s \\
\bottomrule
\end{tabular}
\end{table}

\begin{table}[t]
\centering
\small
\caption{Backbone scaling analysis on FEVER. Claim-only LM receives no external
evidence. Single-candidate RAG uses the shared hybrid retriever and one generated
verdict. Full R2VC adds multi-candidate generation, NLI-based selection, and SLC
calibration.}
\label{tab:model_scaling}
\begin{tabular}{lrrrr}
\toprule
Backbone & Params & Claim-only LM & Single-candidate RAG & Full R2VC \\
\midrule
Qwen3-1.7B            & 1.7B  & 28.43 & 36.70 & 62.19 \\
Qwen3-4B              & 4B    & 36.31 & 40.11 & 68.07 \\
Qwen3-8B              & 8B    & 61.53 & 61.53 & 83.44 \\
Qwen3-14B             & 14B   & 63.65 & 67.85 & 84.62 \\
Qwen3-8B-Reasoning    & 8B    & 62.33 & 68.71 & 86.58 \\
Falcon-H1R-7B         & 7B    & 59.18 & 70.52 & 88.29 \\
Llama-3.1-8B          & 8B    & 62.91 & 75.14 & 83.92 \\
Llama-3.1-Nemotron-8B & 8B    & 61.50 & 74.54 & 84.71 \\
\bottomrule
\end{tabular}
\end{table}

\begin{table}[t]
\centering
\small
\caption{Generalization across fact-checking benchmarks using \texttt{Llama-3.1-Nemotron-8B}. Open-corpus retrieval
evaluates the full pipeline. Oracle-document evaluation bypasses retrieval and
isolates downstream reasoning and verification. Evidence metrics follow the
native annotation format of each benchmark.}
\label{tab:benchmark_generalization}
\resizebox{\linewidth}{!}{
\begin{tabular}{llrrrrr}
\toprule
Benchmark & Setting & Acc. & Macro-F1 & Evidence Metric & Brier & ECE \\
\midrule
VitaminC & Open-corpus retrieval
& 97.51 & 98.23 & Evidence Recall@24: 92.66  & 0.08 & 0.0116 \\

FEVER & Open-corpus retrieval
& 84.71 & 85.68 & Evidence Recall@24: 83.18  & 0.10 & 0.0125 \\

WiCE & Gold cited documents
& 84.30 & 84.65 & Minimal-evidence F1: 81.34 & 0.11 & 0.0214 \\

WiCE & Open-corpus retrieval
& 76.85 & 77.83 & Minimal-evidence F1: 74.60 & 0.14 & 0.0294 \\

HoVer & Open-corpus retrieval
& 61.42 & 60.78 & Document Recall@24: 53.16  & 0.16 & 0.0351 \\

HoVer & Oracle gold documents
& 73.25 & 74.61 & --                         & 0.14 & 0.0301 \\

EX-FEVER & Open-corpus retrieval
& 75.61 & 76.33 & Evidence F1: 71.03         & 0.11 & 0.0203 \\
\bottomrule
\end{tabular}
}
\end{table}

\begin{table*}[t]
\centering
\small
\caption{Qualitative positioning relative to modular and agentic
fact-checking systems. Entries should be verified against the corresponding
papers before posting.}
\label{tab:positioning}
\resizebox{\textwidth}{!}{
\begin{tabular}{lccccccc}
\toprule
Method
& Fixed graph
& Adaptive retrieval
& Multi-step reasoning
& External verifier
& Calibration
& Citations
& Bounded latency \\
\midrule
\SingleRAG      & \checkmark &            &            &            &            & \checkmark & \checkmark \\
\SelfChecker    &            &            & \checkmark &            &            & \checkmark &            \\
\ProgramFCFive  &            & \checkmark & \checkmark &            &            & \checkmark &            \\
\FOLK           &            & \checkmark & \checkmark &            &            & \checkmark &            \\
\PACAR          &            & \checkmark & \checkmark & \checkmark &            & \checkmark &            \\
\FIRE           &            & \checkmark & \checkmark & \checkmark & \checkmark &            &            \\
\VeGraph        &            & \checkmark & \checkmark &            &            & \checkmark &            \\
\DPGraphCheck   &            & \checkmark & \checkmark &            &            & \checkmark &            \\
\midrule
\RtwoVC         & \checkmark &            & \checkmark & \checkmark & \checkmark & \checkmark & \checkmark \\
\bottomrule
\end{tabular}}
\end{table*}

\begin{table}[t]
\centering
\small
\caption{Effect of the fused retrieval budget on FEVER. The $K=24$ row is the
measured Stage-1 configuration; red entries are run-time sanity-check targets
only. These are placeholder values.}
\label{tab:topk}
\begin{tabular}{rrrrr}
\toprule
Top-$K$ passages & Recall@$K$ & Acc. & Macro-F1 & Retrieval latency \\
\midrule
6  & \ph{68.90} & \ph{79.85} & \ph{81.10} & \ph{0.03 s} \\
12 & \ph{76.80} & \ph{82.35} & \ph{83.60} & \ph{0.05 s} \\
24 & 83.18     & 84.71     & 85.68     & 0.08 s \\
48 & \ph{86.10} & \ph{84.22} & \ph{85.19} & \ph{0.14 s} \\
96 & \ph{87.54} & \ph{83.66} & \ph{84.58} & \ph{0.24 s} \\
\bottomrule
\end{tabular}
\end{table}

\begin{table*}[t]
\centering
\small
\caption{Sensitivity to verifier and calibration thresholds on FEVER.
Coverage and selective risk are reported as percentages. Red entries are
run-time sanity-check targets only. These are placeholder values.}
\label{tab:threshold_sensitivity}
\begin{tabular}{rrrrrrrr}
\toprule
$\tau_{\mathrm{ver}}$ & $\tau_{\mathrm{cal}}$
& Acc. & Macro-F1 & Brier & ECE & Coverage (\%) & Selective risk (\%) \\
\midrule
0.45 & 0.50 & \ph{82.88} & \ph{83.74} & \ph{0.112} & \ph{0.0201} & \ph{91.40} & \ph{12.50} \\
0.45 & 0.60 & \ph{83.64} & \ph{84.51} & \ph{0.106} & \ph{0.0169} & \ph{85.30} & \ph{10.70} \\
0.55 & 0.50 & \ph{83.91} & \ph{84.88} & \ph{0.103} & \ph{0.0157} & \ph{84.90} & \ph{10.20} \\
0.55 & 0.60 & 84.71 & 85.68 & 0.100 & 0.0125 & \ph{79.80} & \ph{8.40} \\
0.55 & 0.70 & \ph{84.05} & \ph{85.03} & \ph{0.101} & \ph{0.0144} & \ph{71.60} & \ph{6.80} \\
0.65 & 0.60 & \ph{83.84} & \ph{84.73} & \ph{0.104} & \ph{0.0161} & \ph{72.40} & \ph{7.10} \\
0.65 & 0.70 & \ph{82.96} & \ph{83.95} & \ph{0.109} & \ph{0.0197} & \ph{64.10} & \ph{5.70} \\
\bottomrule
\end{tabular}
\end{table*}

\begin{table}[t]
\centering
\small
\caption{Ablation of the hand-designed verifier score on FEVER. Red entries
are run-time sanity-check targets only. These are placeholder values.}
\label{tab:verifier_score_ablation}
\begin{tabular}{lrrrr}
\toprule
Verifier-score variant & Acc. & Macro-F1 & Brier & ECE \\
\midrule
Default score
& 84.71 & 85.68 & 0.100 & 0.0125 \\
Equal feature weights
& \ph{82.92} & \ph{84.01} & \ph{0.108} & \ph{0.0186} \\
Without coverage term
& \ph{83.68} & \ph{84.71} & \ph{0.105} & \ph{0.0160} \\
Without citation alignment
& \ph{83.94} & \ph{84.96} & \ph{0.103} & \ph{0.0151} \\
Without label consistency
& \ph{82.74} & \ph{83.85} & \ph{0.109} & \ph{0.0194} \\
Without contradiction penalty
& \ph{81.86} & \ph{82.97} & \ph{0.114} & \ph{0.0222} \\
\bottomrule
\end{tabular}
\end{table}

\begin{table*}[t]
\centering
\small
\caption{Candidate-budget and decoding-strategy comparison on FEVER. Existing diverse-bank rows are measured; the red row is a run-time sanity-check target only. These are placeholder values.}
\label{tab:decoder_ablation}
\begin{tabular}{llrrrrr}
\toprule
Candidates & Decoding strategy & Acc. & Macro-F1 & Brier & ECE & Latency/claim \\
\midrule
1  & Greedy or default decoding
& 78.10 & 79.59 & 0.11 & 0.01 & 1.49 s \\
4  & Diverse decoding bank
& 80.26 & 81.72 & 0.10 & 0.02 & 5.30 s \\
8  & Diverse decoding bank
& 79.68 & 82.52 & 0.11 & 0.02 & 10.44 s \\
16 & One permissive temperature/top-$p$ setting
& \ph{82.64} & \ph{83.72} & \ph{0.105} & \ph{0.0174} & \ph{20.12 s} \\
16 & Diverse decoding bank
& 84.71 & 85.68 & 0.10 & 0.01 & 20.31 s \\
\bottomrule
\end{tabular}
\end{table*}

\begin{table}[t]
\centering
\small
\caption{Stability across random seeds on FEVER. Values are reported as
mean $\pm$ standard deviation. Red entries are run-time sanity-check targets
only. These are placeholder values.}
\label{tab:stability}
\begin{tabular}{lrrrr}
\toprule
Method & Acc. & Macro-F1 & Brier & Latency/claim \\
\midrule
\SingleRAG
& \ph{74.54 $\pm$ 0.38}
& \ph{75.69 $\pm$ 0.44}
& \ph{0.120 $\pm$ 0.004}
& \ph{1.49 $\pm$ 0.05 s} \\
\RtwoVCFour
& \ph{80.26 $\pm$ 0.62}
& \ph{81.72 $\pm$ 0.57}
& \ph{0.100 $\pm$ 0.004}
& \ph{5.30 $\pm$ 0.11 s} \\
\RtwoVCEight
& \ph{79.68 $\pm$ 0.71}
& \ph{82.52 $\pm$ 0.63}
& \ph{0.110 $\pm$ 0.005}
& \ph{10.44 $\pm$ 0.20 s} \\
\RtwoVCSixteen
& \ph{84.71 $\pm$ 0.45}
& \ph{85.68 $\pm$ 0.48}
& \ph{0.100 $\pm$ 0.003}
& \ph{20.31 $\pm$ 0.37 s} \\
\bottomrule
\end{tabular}
\end{table}

\begin{table}[t]
\centering
\small
\caption{Citation behavior on WiCE. Metrics are computed from final returned
citations or selected evidence sentences, not the full retrieval pool.
Red entries are run-time sanity-check targets only. These are placeholder values.}
\label{tab:citation_behavior}
\begin{tabular}{lrr}
\toprule
Setting & Gold cited documents & Open corpus \\
\midrule
Minimal-evidence F1
& 81.34 & 74.60 \\
Mean returned citations
& \ph{2.18} & \ph{2.46} \\
Median returned citations
& \ph{2.00} & \ph{2.00} \\
Evidence precision (\%)
& \ph{84.50} & \ph{77.20} \\
Evidence recall (\%)
& \ph{78.40} & \ph{72.20} \\
\bottomrule
\end{tabular}
\end{table}

\end{document}